\documentclass[runningheads]{llncs}

\usepackage{eccv}

\usepackage{eccvabbrv}

\usepackage{graphicx}
\usepackage{booktabs}

\usepackage[accsupp]{axessibility}  % Improves PDF readability for those with disabilities.

\usepackage{hyperref}

\usepackage{orcidlink}

\usepackage{amsfonts,amsmath,amssymb}
\usepackage[dvipsnames]{xcolor}         % colors
\usepackage{subcaption}
\usepackage{float}
\usepackage{graphicx}
\usepackage{wrapfig}
\usepackage{array,multirow}
\usepackage{pifont}

\begin{document}

\title{Pushing the boundaries of event subsampling in event-based video classification using CNNs} 

\titlerunning{Event subsampling in event-based video classification}

\author{Hesam Araghi\orcidlink{0000-0002-4539-4408} \and
Jan van Gemert\orcidlink{0000-0002-3913-2786} \and
Nergis Tomen\orcidlink{0000-0003-3916-1859}}

\authorrunning{H.~Araghi et al.}

\institute{
Computer Vision Lab, Delft University of Technology\\
\email{\{h.araghi, j.c.vangemert, n.tomen\}@tudelft.nl}}

\maketitle

\begin{abstract}
Event cameras offer low-power visual sensing capabilities ideal for edge-device applications. 
However, their high event rate, driven by high temporal details, can be restrictive in terms of bandwidth and computational resources. 
In edge AI applications, determining the minimum amount of events for specific tasks can allow reducing the event rate to improve bandwidth, memory, and processing efficiency.
In this paper, we study the effect of event subsampling on the accuracy of event data classification using convolutional neural network (CNN) models. 
Surprisingly, across various datasets, the number of events per video can be reduced by an order of magnitude with little drop in accuracy, revealing the extent to which we can push the boundaries in accuracy vs. event rate trade-off.
Additionally, we also find that lower classification accuracy in high subsampling rates is not solely attributable to information loss due to the subsampling of the events, but that the training of CNNs can be challenging in highly subsampled scenarios, where the sensitivity to hyperparameters increases. 
We quantify training instability across multiple event-based classification datasets using a novel metric for evaluating the hyperparameter sensitivity of CNNs in different subsampling settings. 
Finally, we analyze the weight gradients of the network to gain insight into this instability.
The code and additional resources for this paper can be found at: \href{https://github.com/hesamaraghi/pushing-boundaries-event-subsampling}{https://github.com/hesamaraghi/pushing-boundaries-event-subsampling}.
  \keywords{Event cameras \and Event-based video classification \and Sparsity of frame representation in event processing \and Accuracy vs. event rate trade-off \and Hyperparameter sensitivity in event data}
\end{abstract}

\section{Introduction}
\label{sec:intro}

Event cameras are power-efficient visual sensors thanks to capturing only \emph{changes} in light intensity.
They combine exceptionally high temporal resolution and low power consumption, making them suitable for edge-device applications~\cite{andersen_event-based_2022,gallego_event-based_2022,hanover_autonomous_2023}. %\jvg{this is important, so indeed you need 3 citations}
Examples include real-time interaction scenarios such as simultaneous localization and mapping (SLAM) in small aerial vehicles~\cite{vidal_ultimate_2018} % optical flow estimation % \jvg each citations take s alot of space in this format; so, because these application are not too importnant, choose just a single citation per application \cite{zhu_unsupervised_2019,hagenaars_self-supervised_2021,paredes-valles_back_2021,shiba_secrets_2022}, 
% \cite{zhu_unsupervised_2019},
and obstacle avoidance 
%\cite{Falanga_Dynamic_2020,walters_evreflex_2021,vemprala_representation_2021}. 
\cite{Falanga_Dynamic_2020}.
%In mobile edge-device applications such as aerial drones, ensuring power efficiency across all components--sensors, data transmission and processing units--is particularly crucial. \jvg{so what?}
In these applications, the visual classification task is essential for identifying, avoiding, or following objects, and navigating through them, for which the convolutional neural network (CNN) is a popular and well-established model. % used for object classification in these settings. 
Considering the total load of the visual pipeline in such edge applications, event cameras have the potential to significantly enhance the overall bandwidth and power efficiency of the system~\cite{censi_power-performance_2015}. However, how to maximize the efficiency gains in the complete framework without compromising task accuracy is not well-studied.

% \jvg{"event-based cameras" or "event cameras"; be consistent; I removed the "based"; as it adds nothing}

In high-motion scenarios, the high temporal resolution of  event cameras generates a large number of events, and the maximum event rate can reach up to one billion events per second %in event cameras with a high number of pixels
\cite{finateu_a_1280_2020,gehrig_are_2022,suh_a_1280_2020}.
Consequently, transmitting and processing such high-rate event data requires substantial bandwidth and computational resources,
reducing the power efficiency of pipelines with event-based sensors in high-motion scenarios.
% thereby increasing power consumption both in data transmission and the processing system.  
% However, from the data transmission point of view event cameras may not be very power-efficient
Increasing the data bandwidth is not always straightforward in edge-device applications, motivating workarounds like filtering out events in hardware \cite{finateu_a_1280_2020} or using event rate control \cite{delbruck_feedback_2021} to prevent the bus saturation.
Another approach is subsampling events %, as
% , effectively reducing computational costs and power consumption.
% the more we subsample events, the more we can 
to reduce the computational cost.
However, subsampling risks information loss and can adversely affect the downstream tasks.
The relationship between subsampling level and its effects on task performance is not straightforward and has been little explored in literature. 
Determining the minimum necessary number of events for particular tasks enables us to tailor the computational processes for edge applications, thereby enhancing the power efficiency and computational effectiveness of the event processing unit.

Here, we study the effect of subsampling on the performance of event data classification using a CNN model.
Surprisingly, our findings show that across multiple datasets, the number of events per video can be reduced by an order of magnitude without sacrificing considerable accuracy.
Moreover, we observe that classification accuracy often remains significantly above chance level even when using as few as 8 or 16 events per video in total (see Figure~\ref{fig:fig_1}).
% ,distributed across multiple frames.
This shows that an unexpected amount of task-relevant information can be retained in just a few events and showcases the potential for greatly reducing the processing power requirements for event-based vision.

At the same time, we uncover unexpected challenges associated with subsampling.
%While class-specific information in the input event stream is reduced by subsampling, resulting in an accuracy drop, 
Specifically, we discover that higher subsampling rates can make the training of CNNs unstable. We explore this subsampling instability and analyze the gradient flow and hyperparameter sensitivity. %Specifically, we show that as sparsity levels increase, a sudden loss in classification accuracy can stem simply from a lack of detailed hyperparameter tuning. Nevertheless, our results suggest that using thorough hyperparameter optimization, the training instability can be alleviated and the task performance can be partially restored.
%Thus, achieving low accuracy is not solely attributable to less information content in the input but also to an increased hyperparameter sensitivity during CNN training in the sparse regime.
 % Although the first trivial effect of the subsampling is the reduction of the information of the input event stream, 
 % which can correlate with a performance drop, 
 % we realized the higher subsampling can make the training of the CNNs unstable. 
 % Thus, the low performance is not merely associated with less information but finding the optimum parameters to train the CNNs with can also be challenging.
%Finally, we study the unstable behavior of the CNN training by performing an analysis of the gradients for the severely subsampled event input. \jvg{I rewrote (simplified; and shortened) it }

The contributions of the paper can be summarized as follows:
\begin{itemize}
    \item 
     We investigate, for the first time, how subsampling affects CNN task performance across various event classification datasets.
    \item We highlight the increased hyperparameter sensitivity in CNN training, particularly at higher subsampling rates.
    \item We introduce a novel metric to quantify the CNN hyperparameter sensitivity in extremely sparse input regimes.
    \item
    We perform a detailed analysis of the gradients to evaluate the effect of subsampling on CNN training stability.
\end{itemize}

% \jvg{All code and data will be shared openly on GitHub.}
% All code and data needed to reproduce the results will be made available in a GitHub repository. %Also the Fan1vs3 dataset will be made public.

\newcolumntype{M}[1]{>{\centering\arraybackslash}m{#1}}
\begin{figure}[t]
\centering
\resizebox{\textwidth}{!}{% <------ Don't
\begin{tabular}{@{}l@{}M{30mm}@{}M{30mm}@{}M{30mm}@{}M{30mm}@{}M{30mm}@{}}
\toprule
\textbf{\small Test Acc.} & \textbf{91.44\%} & \textbf{97.99\%} & \textbf{99.15\%} & \textbf{99.79\%} & \textbf{99.76\%}\\
\textbf{\# events per video} & \textbf{16} & \textbf{32} & \textbf{64} & \textbf{1024} & \textbf{25000}\\
\midrule
{Class `V'} & \includegraphics[width=25mm]{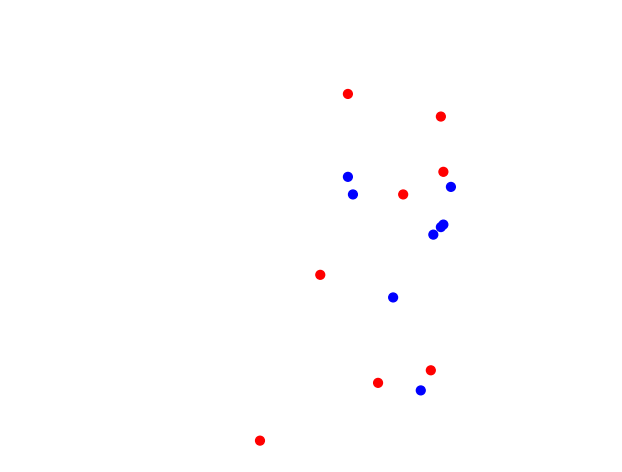} & \includegraphics[width=25mm]{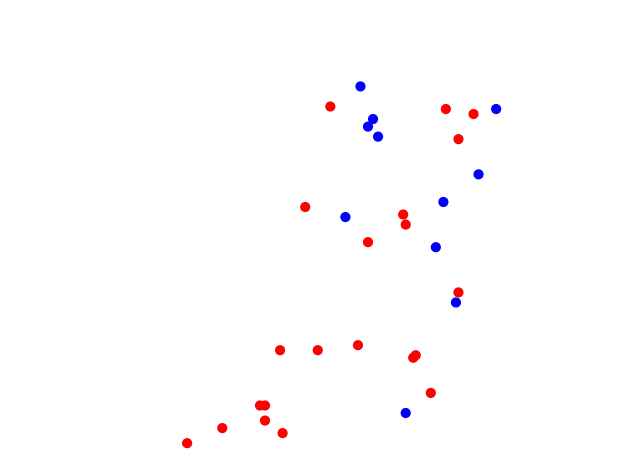} & \includegraphics[width=25mm]{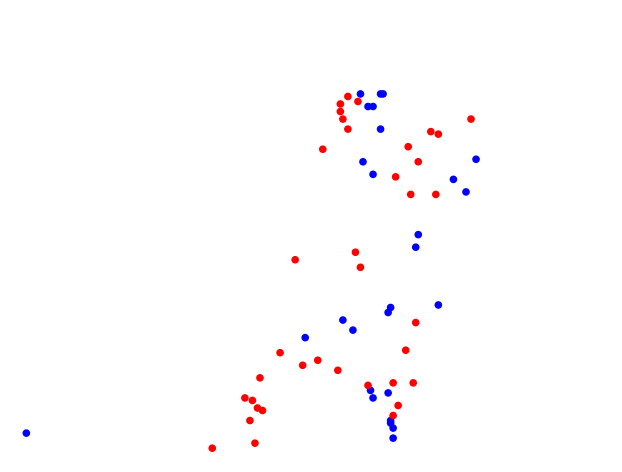} & \includegraphics[width=25mm]{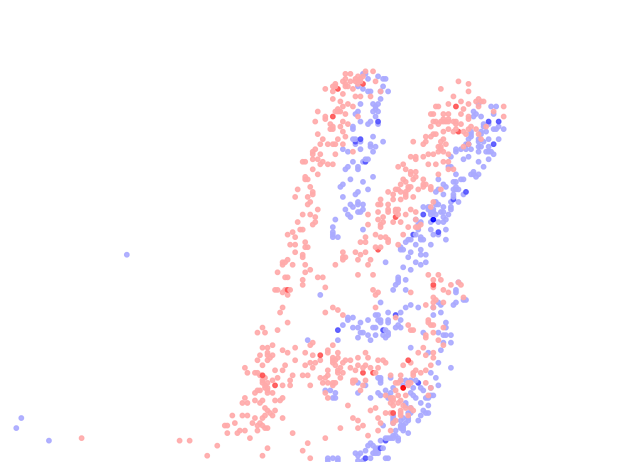} & \includegraphics[width=25mm]{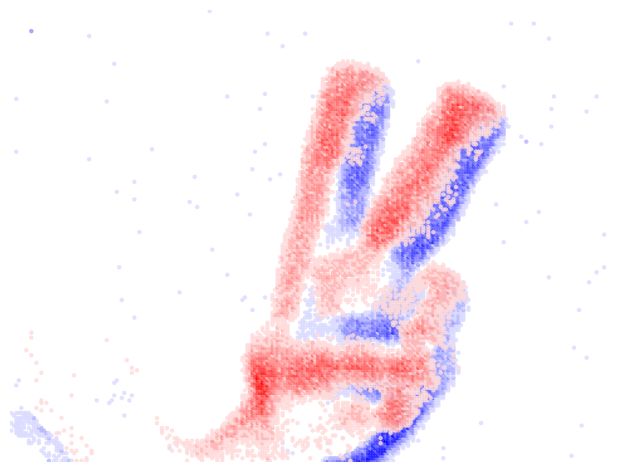}\\
{Class `W'} & \includegraphics[width=25mm]{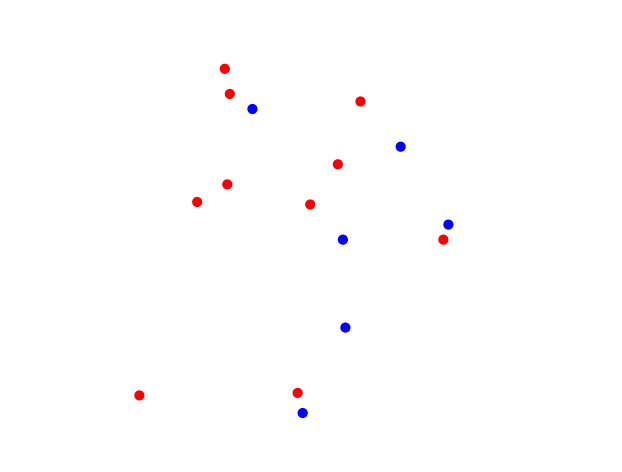} & \includegraphics[width=25mm]{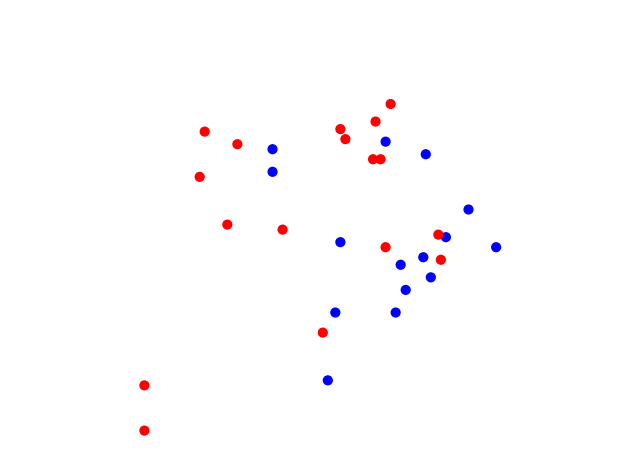} & \includegraphics[width=25mm]{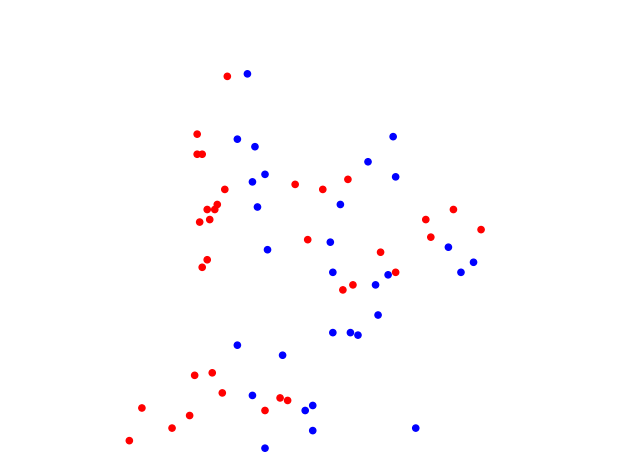} & \includegraphics[width=25mm]{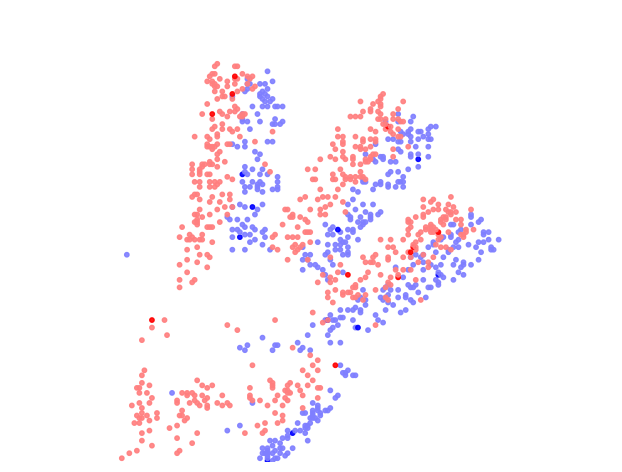} & \includegraphics[width=25mm]{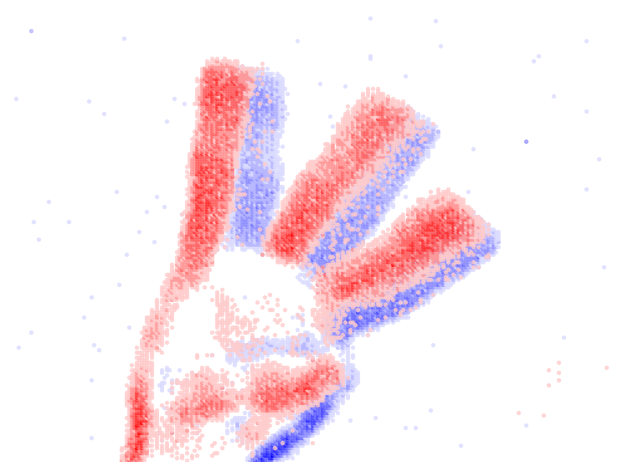}\\
\bottomrule
\end{tabular}
}
\caption{%
Illustration of two classes (`V' and `W') from the Neuromorphic American Sign Language (N-ASL) dataset, which includes 24 classes, for different numbers of events per video. The events are accumulated into a single frame, with red and blue indicating the two polarities. Even with a significantly reduced number of events, a CNN-based classifier can still achieve very high accuracies.}
\label{fig:fig_1}
\end{figure} 

\section{Related work}
\label{sec:related work}

\subsection{Event Classification Using CNNs}
\label{sec:related work subsec:CNNs}

Many tasks in event-based machine vision rely on deep neural networks (DNNs) for processing the events \cite{zheng_deep_2023}.
To use a DNN model, we need to represent the events in a format which is compatible with the network.
One of the commonly used DNNs for event-based machine vision is the convolutional neural network (CNN). 
We concentrate on CNN models in this study since they can offer a balance between task performance and computational efficiency:
While computationally `lighter' methods like handcrafted approaches \cite{lagorce_hots_2017,orchard_hfirst_2015,sekikawa_eventnet_2019} and spiking neural networks \cite{lee_training_2016,shrestha_slayer_2018,vicente-sola_spiking_2023} show limited performance in complex applications, more recent solutions such as visual transformers \cite{klenk_masked_2022,xie_event_2023,zhou_e-clip_2023} come with a high computational cost, diminishing the low power advantages of event cameras during processing.

For processing the events in a CNN, we first need to convert events into frames where events are aggregated into an image-like format. % suitable for CNN-based processing.
Various types of frame-based representations are proposed.
Time surface representation is an image, with pixel values representing timestamps.
For instance, \cite{lagorce_hots_2017} stores the timestamp of the most recent event for each pixel.
% To mitigate noise sensitivity, \cite{sironi_hats_2018} takes the average of event timestamps in neighboring spatiotemporal space.
Time surface enables asynchronous updates and retains temporal information of last events.
Event count/histogram is another image representation generated by counting events in a specific time window for each pixel \cite{maqueda_event-based_2018}. 
% In addition to event counts, \cite{zhu_ev-flownet_2018} also stores the timestamps of the last events.
This representation can have the spatial information of the edges, but it may result in blurry outputs for fast-moving objects.
Another approach is the voxel grid representation \cite{zhu_unsupervised_2019}, which discretizes the time window into multiple bins and computes an image for each bin through a weighted summation of events near that bin. 
 In contrast to predefined representations, \cite{gehrig_end--end_2019} proposed the EST algorithm, which is a learning-based method to generate a voxel grid representation which can be trained flexibly for various tasks. 
 % Being fully differentiable, this trainable representation allows for maximizing performance on a specific task.
Due to its end-to-end representation learning property, we use the EST algorithm~\cite{gehrig_end--end_2019} to assess the classification performance of CNNs across different levels of sparsity.
We specifically investigate the effects of input \emph{sparsity} on training and classification accuracy, since input frames constructed from subsampled events lead to increased sparsity and thus to more efficient processing.

\subsection{Sparsity in Event Cameras}
\label{sec:related work subsec:sparsity}

%  \jvg{less is more} the sparsity properties of event data have focused on benefiting this 
Previous works considering sparsity to improve computational efficiency in event processing include modeling events as point-clouds \cite{sekikawa_eventnet_2019}, graphs \cite{gehrig_pushing_2022}, and employing sparse CNNs \cite{vedaldi_event-based_2020}.
These models can reduce the computational costs by increasing the sparsity.
However, in this work, we focus on accuracy-sparsity trade-off instead of computational efficiency.
% There are also studies aimed at mitigating the impact of input data sparsity on method performance \cite{uhrig_sparsity_2017, jaritz_sparse_2018, huang_hms-net_2020}. 
% These studies have primarily addressed depth estimation from sparse lidar data, where the adaptability of these methods to event camera inputs is uncertain.
To reduce the bandwidth and computation costs, \cite{fischer_how_2022} proposes using a small subset of pixels generating events for visual place recognition. 
However, this method relies on event counting for selected pixels and may not be suitable for other tasks like image classification.
In \cite{gruel_event_2022}, the authors introduce spatial downscaling methods based on either averaging events or estimating luminance. 
Similarly, in \cite{gruel_performance_2023}, two additional spatial downscaling methods are proposed, using SNN pooling. 
These techniques effectively reduce the number of events, and their effect on classification accuracy using SNNs is studied in \cite{gruel_performance_2023, gruel_frugal_2023}.
The work by \cite{cohen_spatial_2018} examines the effect of event downsampling on event rate reduction and classification accuracy, suggesting that reducing the spatial and temporal resolution of input data can improve classification accuracy and lower data rates while retaining the number of events.
They employ a synaptic kernel inverse method (SKIM) \cite{tapson_synthesis_2013}.
However, the performance of SKIM and SNN-based methods used in \cite{cohen_spatial_2018, gruel_performance_2023, gruel_frugal_2023} does not match state-of-the-art methods, raising questions about the generalizability of their findings to more accurate models like CNNs. 
In contrast, our work focuses on the impact of event subsampling, where the number of events is significantly reduced. 
We examine the effect of subsampling across multiple event datasets and introduce the challenges of training networks with severe subsampling by studying hyperparameter sensitivity.

% The work by \cite{cohen_spatial_2018} is the most closely related to ours. 
% They examine the effect of event downsampling on event rate reduction and classification accuracy, suggesting that reducing the spatial and temporal resolution of input data can improve classification accuracy and lower data rates. 
% Their method retains the number of events while reducing spatial and temporal resolution.
% They employ a synaptic kernel inverse method (SKIM) \cite{tapson_synthesis_2013}, where its performance does not match state-of-the-art methods, raising questions about the generalizability of their findings to more accurate models like CNNs.
% In contrast, our work focuses on the impact of event subsampling, where the number of events reduces significantly. 
% We also introduce the challenges of training networks with severe subsampling by studying hyperparameter sensitivity.
\section{Method}
\label{sec:method}

% This section provides a brief overview of the Event Spike Tensor (EST) algorithm \cite{gehrig_end--end_2019} as our CNN-based method for event-based video classification, along with the details of the training procedure.
% Additionally, we present the datasets used in the experiment of this paper.

\subsection{The EST Algorithm for Converting Events to Frames}
\label{sec:method:subsec:EST overview}

%To use CNNs with event data, the events need to be converted into grid-like representations suitable for CNNs. 
The EST algorithm \cite{gehrig_end--end_2019} automatically learns the mapping between the events and the frames from the data, making it adaptable to specific tasks. 
Suppose we have a set of events in a video denoted by $\mathcal{E}=\{e_i\}_{i=1}^N$, where $e_i=(x_i,y_i,t_i,p_i)$ contains the spatial position of the event $(x_i,y_i)$, occurring time $t_i$, and the polarity of the event $p_i\in\{-1,+1\}$.
The temporal dimension of the video is divided into $C$ equally spaced timestamps $\{t^{(c)}\}_{c=1}^C$, and for each timestamp there are two frames one for each polarity, resulting in total $2 C$ frames.
% We can show the frame representation by a tensor $\mathbf{V}\in\mathbb{R}^{2\times C \times H \times W}$ 
The algorithm trains a multilayer perceptions (MLP) layer as a filter function $f_{\theta}(\cdot): \mathbb{R}\rightarrow\mathbb{R}$ with learnable parameters $\theta$  to compute the frame representation $V$.
Particularly, the value at position $(x,y)$ of frame $c$ with polarity $p\in\{0,1\}$ equals
\begin{equation}
V(p,c,x,y) = \sum_{i=1}^{N} t_i\,f_{\theta}\left(t_i - t^{(c)}\right)\,\mathcal{I}(x_i - x)\mathcal{I}(y_i - y)\mathcal{I}(p_i - p),
\label{eq:aggregate_timestamps}
\end{equation}
where $\mathcal{I}(\cdot):\mathbb{R}\rightarrow \{0,1\}$ is an indicator function which outputs 1 for zero input and 0 otherwise.
After generating the $2 C$ frames from $N$ number of events, a CNN model performs the classification task, where the number of input channels for the CNN model equals the number of frames, i.e. $2 C$.

\subsection{Event Subsampling Procedure}
\label{sec:method:subsec:subsampling_procedure}
% \textbf{(Nergis:How we subsample, why we subsample again at every epoch.
% Justify why aren't we subsampling spatially/temporally only?
% Maybe not a whole subsection, but please make sure the reader is clear on the subsampling strategy + justification.
% Also, add under what conditions this training strategy makes sense in discussion!}

For event subsampling, we randomly select from the events within a video.
During training, in every epoch, we subsample a new subset of events for each video rather than fixing the subsampled events for each video during the entire training process.
This approach helps to mitigate overfitting on the training set, which could occur if a fixed subset of events is used in each epoch.
It's particularly important in scenarios with sparse input, where the subsampling rate is high. 
Additionally, we do not restrict the subsampling of events to specific spatial pixels or time intervals. 
Random subsampling provides better augmentation of the videos during training, improving the model's robustness to different subsets of events in a video. 
This provides better generalization to sparse inputs at test time, which might be crucial for real-time applications.
By introducing different subsets of events to the network in every epoch, we try to minimize the impact of information loss and isolate the effect of sparsity on classification accuracy.
The practical implications of this training approach are discussed in Section~\ref{sec:discussion}. %\jvg{say something about test-time sub-sampling}
During testing, the evaluation is conducted 20 times with different random subsamples for the test set, and the average accuracy is reported. 

\subsection{Training Procedure}
\label{sec:method:subsec:training_procedure}

To convert the events into frames, we adopt the settings from the original EST paper \cite{gehrig_end--end_2019} regarding the number of frames. 
We choose $C = 9$ frames for each positive and negative polarity for any video, yielding a total of 18 channels for the input of the CNN classifier. 
As in Equation~\eqref{eq:aggregate_timestamps}, 
% each event $e_i$ influences only the values of the pixels in the frames which share the same spatial position, $(x_i,y_i)$, as the event. 
% Consequently, 
pixels that do not correspond to any event's spatial location are assigned zero values. 
Thus, in this conversion method, higher subsampling rates of the events in the video lead to increased sparsity of the input frames for the CNN. 
Therefore, studying the effect of subsampling is similar to studying the impact of sparsity. 
Throughout the paper, we train a ResNet34~\cite{he_deep_2015} as the CNN classifier, using the training set of the corresponding datasets. Before training, the weights are always initialized from a ResNet34 model pre-trained on ImageNet1k\_v1.
% See supplementary material for further training details. 
More details of the EST algorithm are in the supplementary material. %TODO: supp

\subsection{Event Classification Datasets}
% TODO: Samples of each in supp material (or only Fan dataset)
%TODO:  train val test split for each dataset
%TODO: image resolutions
% TODO: table with number of events and classes and number of videos

\subsubsection{N-Caltech101 \cite{orchard_converting_2015}:} 
This dataset is derived from Caltech101~\cite{fei_learning_2004}, which comprises 101 categories of images of various objects. 
To capture events from the images, \cite{orchard_converting_2015} employed an ATIS event camera \cite{posch_atis_2010} mounted in front of an LCD monitor displaying the images.
The camera was then moved in three directions to generate events.
% The dataset comprises 8,709 videos, each with an approximate duration of 300 milliseconds.
% However, this dataset may not be optimal for event classification tasks, since the temporal details of the events in this dataset may not be very informative for classifying the objects.
The camera motion pattern is identical for all classes, meaning that the temporal information derived from the camera's movement may not provide useful features for distinguishing between different object classes.
%\vspace{-2ex}

\subsubsection{N-Cars \cite{sironi_hats_2018}:} 
This dataset captures real-world scenes by mounting an ATIS camera behind the windshield of a car and recording videos while driving in urban environments. 
It contains two classes of cars and background scenes, featuring various background scenarios.
% Each video has a duration of approximately 100 milliseconds. 
Unlike N-Caltech101, motion has a stronger correlations with the class labels.
%\vspace{-2ex}

\subsubsection{N-ASL \cite{bi_graph-based_2019}:} 
This dataset consists of handshape movements recorded by a DAVIS240c event camera, featuring 24 classes corresponding to 24 letters (excluding J and Z) from American Sign Language (ASL).
 The camera was stationary while subjects performed the handshape of each letter.
% There are 4,200 videos for each letter, resulting in a total of 100,800 videos. 
% Each video has a duration of approximately 100 milliseconds. 
%This is also a real-world dataset and has more classes than N-Cars. 
%However, due to the relatively simple backgrounds, this dataset is not particularly challenging for classification tasks.
%\vspace{-2ex}

\subsubsection{DVS-Gesture \cite{amir_low_2017}:} 
This dataset demonstrates 11 classes of hand and arm gestures from 29 subjects under 3 different lighting conditions.
The gestures were performed in front of a DVS128 event camera against a stationary background.
Some examples of gestures are hand waving, arm rotating, air guitar, and an ``other" gesture invented by the subject.
We used the Tonic library \cite{lenz_tonic_2021} for downloading and loading the dataset.
%\vspace{-2ex}

\subsubsection{Fan1vs3:}  We introduce a new dataset to evaluate motion to better address the microsecond temporal resolution of event cameras.
We placed a Prophesee Gen4 event camera \cite{finateu_a_1280_2020} in front of a rotating fan, with the blades set to two different speeds, slow (level 1) vs. fast (level 3), corresponding to the two classes of the dataset.
Distinguishing between the classes in this dataset relies more on the temporal details of the events rather than their spatial information, which makes it different than the previously mentioned datasets, where spatial information plays a more important role in classifying the objects. 
Further details of this toy dataset are provided in the supplementary material. % TODO: supp
% To push the importance of temporal details and better utilize the high temporal details of events 
%\vspace{-2ex}

\section{Experiments}
\label{sec:experiments}

\subsection{Accuracy vs. Sparsity in Event Data Classification}
\label{sec:experiments subsec:accuracy-sparsity}

We test the resilience of classification accuracy against reducing number of events per video across various datasets, including \textbf{N-Caltech101} \cite{orchard_converting_2015}, \textbf{N-ASL} \cite{bi_graph-based_2019}, \textbf{N-Cars} \cite{sironi_hats_2018}, \textbf{DVS-Gesture} \cite{amir_low_2017}, and the proposed \textbf{Fan1vs3}.
% \footnote{The code used to reproduce the results will be made available in a GitHub repository. Additionally, the Fan1vs3 toy dataset will be accessible online.}
We use Adam \cite{kingma_adam_2017} for optimization with the `Reduce on Plateau' learning rate scheduler for all experiments.
The reduction factor is set at 2, and the patience parameter is determined based on the number of epochs, with values of 20 for 100 epochs, 40 for 250 epochs, and 75 for 500 epochs. 
The batch size, learning rate, weight decay coefficient, and number of epochs are chosen to be the same across all sparsity levels of the same dataset. For each dataset, we picked a set of hyperparameters which could converge to a high accuracy at every sparsity level based on a small preliminary experiments, except N-Cars, for which we adopt the values from the EST paper \cite{gehrig_end--end_2019}. For the exact values of the hyperparameters, please see the supplementary material. %TODO 
For the test set, the evaluation is conducted 20 times with different random subsamples, and the average accuracy is reported.

\begin{table}[tb]
    \centering
    \caption{%
    CNN classification accuracy for decreasing the number of events per video across various datasets. We can reduce the number of events per video by an order of magnitude without sacrificing considerable accuracy.
    Even under extreme sparsity (8 or 16 total events per video with 18 frames) the accuracies remain significantly above chance level.%
    % , remaining significantly above the chance levels even when using as few as 8 or 16 events per video in total,
    % This table illustrates that there is significant potential to increase the subsampling rates for improving the power efficiency of data transmission and processing while maintaining satisfactory accuracy for the application and reducing bandwidth requirements.
    % Across all datasets except Fan1vs3, the $p$-values are almost zero for various subsampling levels, indicating that even with a very small number of events, such as 8 or 16 events, the accuracies are significantly above the chance level. (Nergis: I commented this out, as you already say this in the text.)
    % For the FAN1vs3 dataset, the temporal details of events are more important than spatial details.
    %  Consequently, the network encounters challenges in classifying speeds in high sparsity levels of the input, exemplifying an extreme case where CNNs struggle with input sparsity.%
    }
    \label{tab:sparsity_vs_acc}
        \resizebox{\textwidth}{!}{% <------ Don't forget this %
    \begin{tabular}{ccccccccccc}
\toprule
 & & & \multicolumn{8}{c}{\# events per video}\\
Dataset & \# classes & & 8 & 16 & 32 & 64 & 512 & 1024 & 4096 & 25000\\
\midrule
N-ASL & 24 & Test Acc. (\%) & $24.33$ & $91.44$ & $97.99$ & $99.15$ & $99.80$ & $99.79$ & $99.81$ & $99.76$ \\
 & & \textcolor{WildStrawberry}{\scriptsize Std. Dev. (\%)} & \textcolor{WildStrawberry}{\scriptsize$\pm 28.52$} & \textcolor{WildStrawberry}{\scriptsize$\pm 0.37$} & \textcolor{WildStrawberry}{\scriptsize$\pm 0.23$} & \textcolor{WildStrawberry}{\scriptsize$\pm 0.08$} & \textcolor{WildStrawberry}{\scriptsize$\pm 0.01$} & \textcolor{WildStrawberry}{\scriptsize$\pm 0.09$} & \textcolor{WildStrawberry}{\scriptsize$\pm 0.15$} & \textcolor{WildStrawberry}{\scriptsize$\pm 0.28$} \\
 & & \textcolor{Cerulean}{\scriptsize p-value} & \textcolor{Cerulean}{\scriptsize 0} & \textcolor{Cerulean}{\scriptsize 0} & \textcolor{Cerulean}{\scriptsize 0} & \textcolor{Cerulean}{\scriptsize 0} & \textcolor{Cerulean}{\scriptsize 0} & \textcolor{Cerulean}{\scriptsize 0} & \textcolor{Cerulean}{\scriptsize 0} & \textcolor{Cerulean}{\scriptsize 0} \\
N-Cars & 2 & Test Acc. (\%) & $72.51$ & $78.74$ & $83.23$ & $86.58$ & $92.97$ & $93.23$ & $92.46$ & $91.87$ \\
 & & \textcolor{WildStrawberry}{\scriptsize Std. Dev. (\%)} & \textcolor{WildStrawberry}{\scriptsize$\pm 0.18$} & \textcolor{WildStrawberry}{\scriptsize$\pm 0.12$} & \textcolor{WildStrawberry}{\scriptsize$\pm 0.17$} & \textcolor{WildStrawberry}{\scriptsize$\pm 0.16$} & \textcolor{WildStrawberry}{\scriptsize$\pm 0.13$} & \textcolor{WildStrawberry}{\scriptsize$\pm 0.20$} & \textcolor{WildStrawberry}{\scriptsize$\pm 0.95$} & \textcolor{WildStrawberry}{\scriptsize$\pm 0.81$} \\
 & & \textcolor{Cerulean}{\scriptsize p-value} & \textcolor{Cerulean}{\scriptsize 0} & \textcolor{Cerulean}{\scriptsize 0} & \textcolor{Cerulean}{\scriptsize 0} & \textcolor{Cerulean}{\scriptsize 0} & \textcolor{Cerulean}{\scriptsize 0} & \textcolor{Cerulean}{\scriptsize 0} & \textcolor{Cerulean}{\scriptsize 0} & \textcolor{Cerulean}{\scriptsize 0} \\
DVS-Gesture & 11 & Test Acc. (\%) & $47.98$ & $55.99$ & $75.23$ & $84.81$ & $93.37$ & $94.69$ & $95.18$ & $95.33$ \\
 & & \textcolor{WildStrawberry}{\scriptsize Std. Dev. (\%)} & \textcolor{WildStrawberry}{\scriptsize$\pm 0.21$} & \textcolor{WildStrawberry}{\scriptsize$\pm 0.23$} & \textcolor{WildStrawberry}{\scriptsize$\pm 2.21$} & \textcolor{WildStrawberry}{\scriptsize$\pm 0.34$} & \textcolor{WildStrawberry}{\scriptsize$\pm 0.65$} & \textcolor{WildStrawberry}{\scriptsize$\pm 0.30$} & \textcolor{WildStrawberry}{\scriptsize$\pm 0.49$} & \textcolor{WildStrawberry}{\scriptsize$\pm 0.52$} \\
 & & \textcolor{Cerulean}{\scriptsize p-value} & \textcolor{Cerulean}{\scriptsize 1.08e-205} & \textcolor{Cerulean}{\scriptsize 3.25e-282} & \textcolor{Cerulean}{\scriptsize 0} & \textcolor{Cerulean}{\scriptsize 0} & \textcolor{Cerulean}{\scriptsize 0} & \textcolor{Cerulean}{\scriptsize 0} & \textcolor{Cerulean}{\scriptsize 0} & \textcolor{Cerulean}{\scriptsize 0} \\
Fan1vs3 & 2 & Test Acc. (\%) & $53.35$ & $54.17$ & $56.62$ & $58.53$ & $75.29$ & $94.00$ & $98.24$ & $99.40$ \\
 & & \textcolor{WildStrawberry}{\scriptsize Std. Dev. (\%)} & \textcolor{WildStrawberry}{\scriptsize$\pm 0.56$} & \textcolor{WildStrawberry}{\scriptsize$\pm 1.77$} & \textcolor{WildStrawberry}{\scriptsize$\pm 0.42$} & \textcolor{WildStrawberry}{\scriptsize$\pm 0.99$} & \textcolor{WildStrawberry}{\scriptsize$\pm 1.19$} & \textcolor{WildStrawberry}{\scriptsize$\pm 5.69$} & \textcolor{WildStrawberry}{\scriptsize$\pm 0.79$} & \textcolor{WildStrawberry}{\scriptsize$\pm 0.51$} \\
 & & \textcolor{Cerulean}{\scriptsize p-value} & \textcolor{Cerulean}{\scriptsize 3.67e-01} & \textcolor{Cerulean}{\scriptsize 2.86e-01} & \textcolor{Cerulean}{\scriptsize 1.54e-01} & \textcolor{Cerulean}{\scriptsize 1.06e-01} & \textcolor{Cerulean}{\scriptsize 9.76e-06} & \textcolor{Cerulean}{\scriptsize 7.48e-17} & \textcolor{Cerulean}{\scriptsize 1.02e-20} & \textcolor{Cerulean}{\scriptsize 2.61e-22} \\
N-Caltech101 & 101 & Test Acc. (\%) & $25.20$ & $31.37$ & $39.00$ & $46.19$ & $67.85$ & $74.06$ & $82.87$ & $88.62$ \\
 & & \textcolor{WildStrawberry}{\scriptsize Std. Dev. (\%)} & \textcolor{WildStrawberry}{\scriptsize$\pm 0.17$} & \textcolor{WildStrawberry}{\scriptsize$\pm 0.15$} & \textcolor{WildStrawberry}{\scriptsize$\pm 0.21$} & \textcolor{WildStrawberry}{\scriptsize$\pm 0.19$} & \textcolor{WildStrawberry}{\scriptsize$\pm 0.42$} & \textcolor{WildStrawberry}{\scriptsize$\pm 0.54$} & \textcolor{WildStrawberry}{\scriptsize$\pm 0.47$} & \textcolor{WildStrawberry}{\scriptsize$\pm 0.51$} \\
 & & \textcolor{Cerulean}{\scriptsize p-value} & \textcolor{Cerulean}{\scriptsize 0} & \textcolor{Cerulean}{\scriptsize 0} & \textcolor{Cerulean}{\scriptsize 0} & \textcolor{Cerulean}{\scriptsize 0} & \textcolor{Cerulean}{\scriptsize 0} & \textcolor{Cerulean}{\scriptsize 0} & \textcolor{Cerulean}{\scriptsize 0} & \textcolor{Cerulean}{\scriptsize 0} \\
\bottomrule
\end{tabular}
% <------ Don't forget this %
    }
\end{table}

In \cref{tab:sparsity_vs_acc}, we present the classification accuracy of the EST algorithm with the ResNet34 model with varying numbers of subsampling for different datasets. 
Each entry represents the average accuracy over 5 independent runs with different seeds  (in black), and their standard deviation (in red).
The results demonstrate a decrease in accuracy as the number of events per video decreases, which is not surprising due to the loss of information incurred by removing events.
% , thereby making it more challenging for the network to differentiate between classes.
However, it's notable that the network maintains accuracy levels close to those achieved with denser cases for even unexpectedly small numbers of events per video.
For instance, in the N-ASL dataset, even with only 64 events per video, distributed over 18 frames, the accuracy remains at 99 percent.
Similarly, across other datasets, we observe that the CNN can still perform classification above chance level while reducing total events per video considerably.
To show that the accuracies significantly exceed the corresponding chance levels, we compute the $p$-value for each mean accuracy in \cref{tab:sparsity_vs_acc} using the one-tailed binomial test (in blue).
For all datasets, except Fan1vs3, the $p$-values are almost zero for different subsampling levels.
Thus, even with just 8 or 16 events per video, distributed over 18 frames, the accuracies are significantly higher than chance level.
Here, the Fan1vs3 dataset serves as an extreme case where CNNs struggle with the high sparsity of the input.
Because the classes are distinguished by the fan speed, the temporal details of the events play a more crucial role than the spatial details.
Our procedure of randomly subsampling the events then makes it challenging for the network to accurately classify speeds, especially as the number of events is decreased.
This highlights the subsampling sensitivity in applications where temporal details are more important because as we reduce the number of events, temporal information is lost faster than spatial information.
For better visualization, the accuracy curves from \cref{tab:sparsity_vs_acc}, along with their standard deviations, are presented in the supplementary material.

% \begin{table}
%     \centering
%     \caption{%
% Comparing the $p$-value for the null hypothesis that `accuracies are at chance level' using the binomial test. 
% Across all datasets except FAN1vs3, the $p$-values are almost zero for various subsampling levels.
%  This indicates that even with a very small number of events, such as 8 or 16 events, the accuracies are significantly above the chance level.
% For the FAN1vs3 dataset, where the classes are differentiated based on the speed of the fan, the temporal details of events are more important than spatial details.
%  Consequently, the network encounters challenges in classifying speeds in high sparsity levels of the input. 
%  The FAN1vs3 dataset exemplifies an extreme case where CNNs struggle with input sparsity.
%     }
%     \label{tab:sparsity_vs_acc}
%       \resizebox{0.7\textwidth}{!}{% <------ Don't forget this %
%     \input{images/sparsity_vs_acc/p_values}% <------ Don't forget this %
%     }
% \end{table}

\subsection{Hyperparameter Sensitivity with Increasing Sparsity}
\label{sec:experiments subsec: HPO}

In this subsection, we study the effect of subsampling on the \emph{training procedure of a CNN model}.
While experiments of the previous subsection have shown that CNNs can maintain high accuracies even with sparse input data, it is essential to recognize that sparse input in CNNs introduces new challenges. 
% Here, we argue that training CNNs in sparse cases may pose greater challenges.
As we sample less events per video, we inevitably lose information from the removed events, and the trivial consequence is a drop in classification accuracy.
 % However, the loss of information is not the only contributing factor to the accuracy drop.
However, in addition to this effect, we find that the sensitivity of training to hyperparameters may also increase, making the training more challenging.
To investigate it further, we compare the distributions of classification accuracy for different optimization hyperparameter sets in two scenarios: 1.~with sparse input and 2.~with dense input.

In our analysis, we focus on four datasets --- Fan1vs3, DVS-Gesture, N\-Caltech101, and N-ASL --- to analyze how varying levels of subsampling affect the training process and sensitivity to optimization hyperparameters.
We select three optimization hyperparameters (HPs) to tune: learning rate, batch size, and weight decay.
 To cover a wide range of possibilities, we choose learning rates from the set $\{10^{-i}\}_{i=2}^6$, ensuring that the range is large enough to have performance drops at both the upper and lower bounds. 
Weight decay values are selected from the set $\{0\} \cup \{10^{-i}\}_{i=1}^{4}$. 
For batch sizes, we consider $\{1, 8, 16, 32, 64, 128\}$ for all datasets except Fan1vs3, where we exclude a batch size of 128 due to GPU memory limitations.
We determine the number of epochs such that the loss curves reach convergence. 
Specifically, we choose 250 epochs for Fan1vs3, DVS-Gesture, and N-Caltech101, and 100 epochs for N-ASL. 
Other HPs, such as the optimizer (Adam), learning rate scheduler, reduce factor, patience, and number of epochs remain consistent with those used in \cref{sec:experiments subsec:accuracy-sparsity}.
We randomly select 50 sets of HPs from the specified settings. 
For each set, we conduct training with 3 different seeds for both sparse and dense cases, resulting in a total of 300 different training instances per dataset.
In the dense case, we choose 25,000 events per video, while for the sparse case, we select 1,024 events per video for Fan1vs3, 64  for DVS-Gesture, 512 for N-Caltech101, and 8 for N-ASL. 
The dataset-specific choice of the number of events per video in sparse case is based on the sparsity levels where we started observing instability in training for each dataset. 
For example, for N-ASL, the number of events per video had to be decreased to below 16 to observe a substantial drop in accuracy.

\begin{figure}[tb]
\centering
    \begin{subfigure}[b]{0.8\textwidth}
\includegraphics[width=1\textwidth]{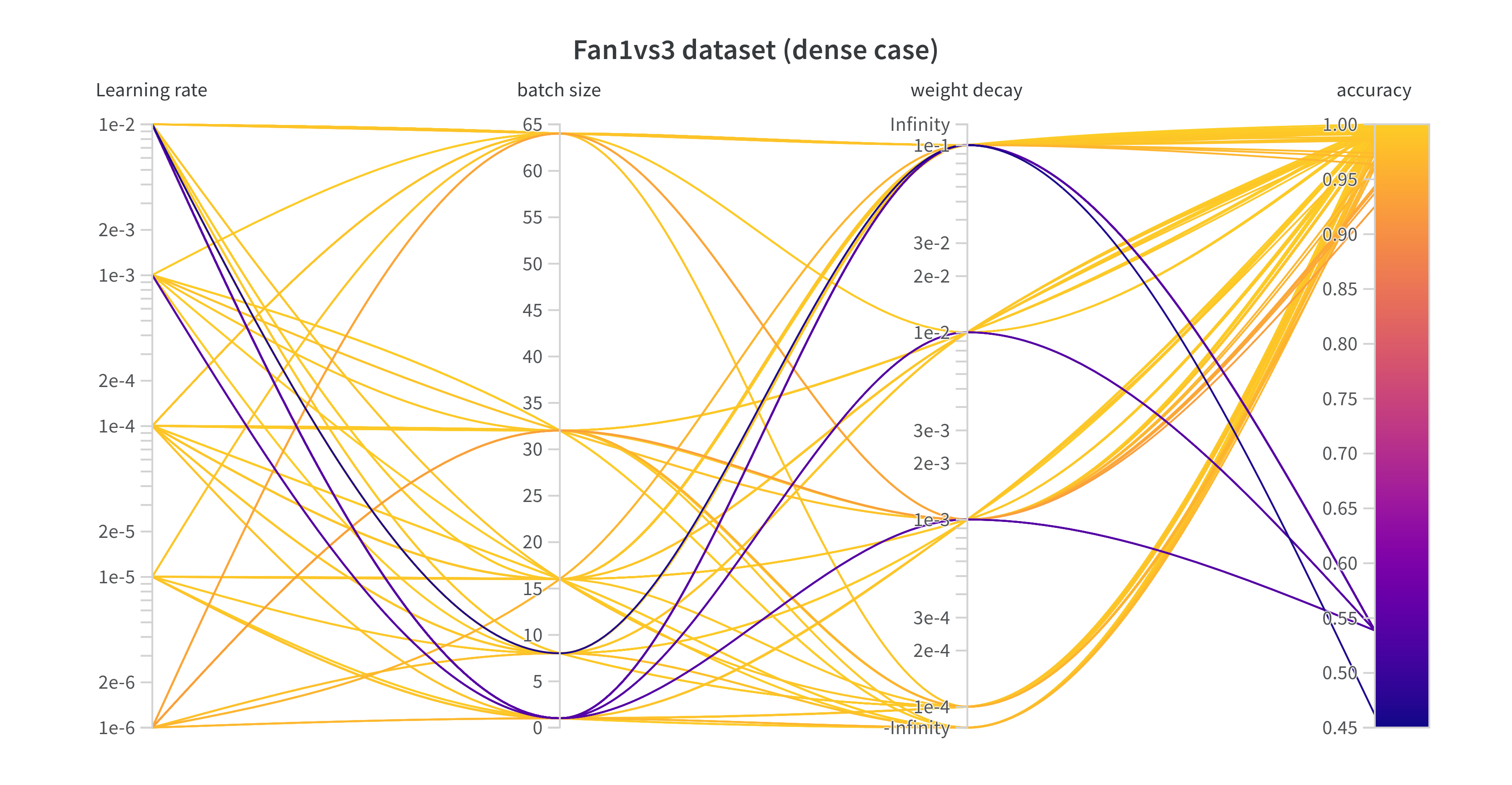}
        \caption{Fan1vs3 (dense input with 25,000 events)}
        \label{fig:fan1vs3_string_dense}
    \end{subfigure}
    \\
    \begin{subfigure}[b]{0.8\textwidth}
\includegraphics[width=1\textwidth]{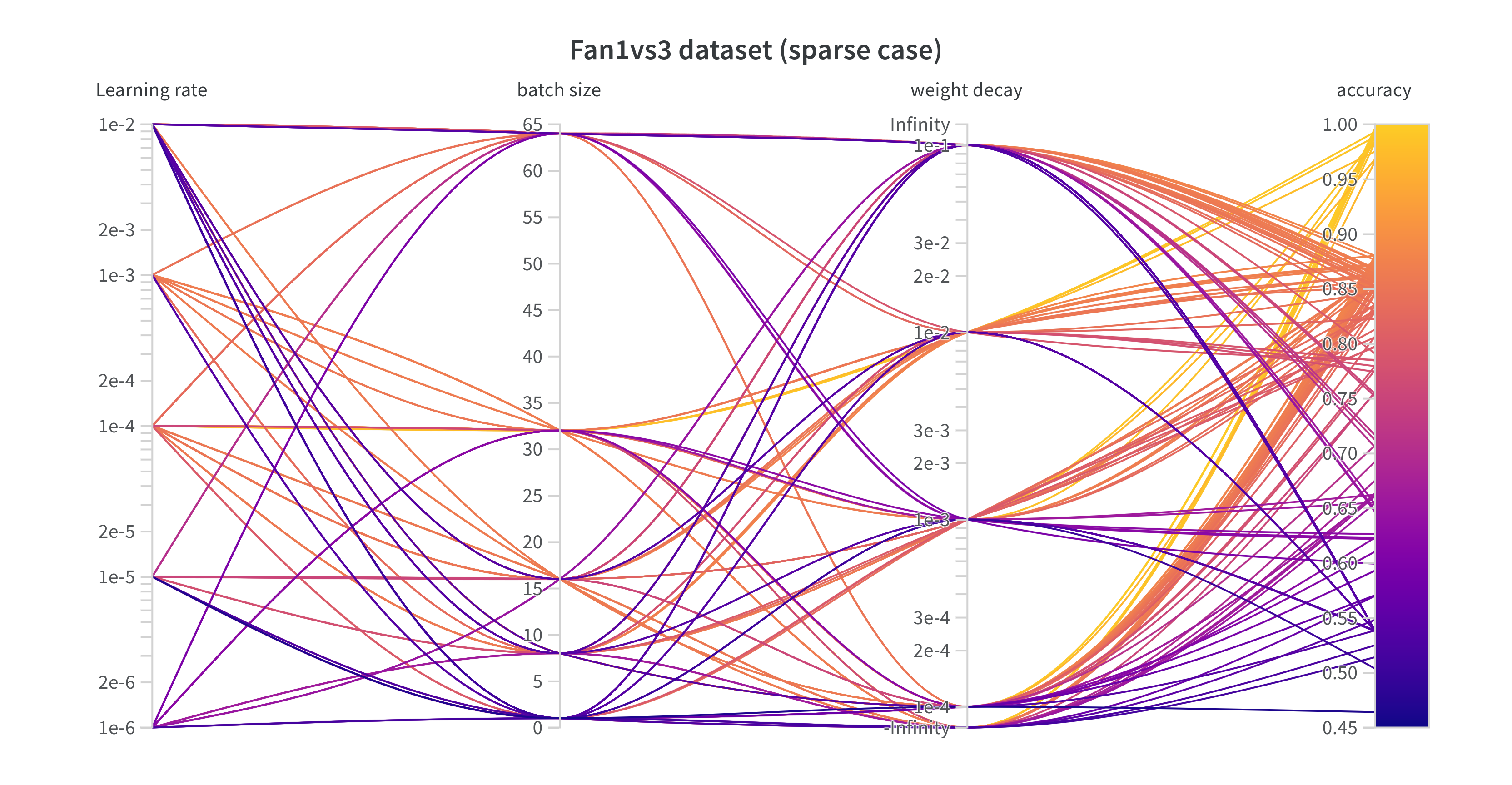}
        \caption{Fan1vs3 (sparse input with 1,024 events)}
        \label{fig:fan1vs3_string_sparse}
    \end{subfigure}
    \caption{Parallel coordinate plots showing HP tuning results for Fan1vs3 dataset using dense and sparse inputs. HPs: learning rate, batch size, and weight decay. In the dense setting, we observe test accuracies concentrated near the maximum accuracy, while in the sparse setting, we observe a small number of runs achieving the maximum accuracy. The plots are from the Weight and Biases website \cite{wandb}.
    }
    \label{fig:HP tuning strings}
\end{figure}

\begin{figure}[tb]
\centering
\resizebox{\textwidth}{!}{% <------ Don't
\begin{tabular}{@{}l@{}M{50mm}@{}M{30mm}@{}M{30mm}@{}M{30mm}@{}}
\toprule
 & {\scriptsize Hyperparameter Tuning Histogram} & {\scriptsize ~~~~~~Learning rate vs Acc.} & {\scriptsize ~~~~~Batch size vs Acc.} & {\scriptsize ~~~~~Weight decay vs Acc.}\\
\midrule
\rotatebox{90}{\scriptsize Fan1vs3} & {\includegraphics[width=50mm]{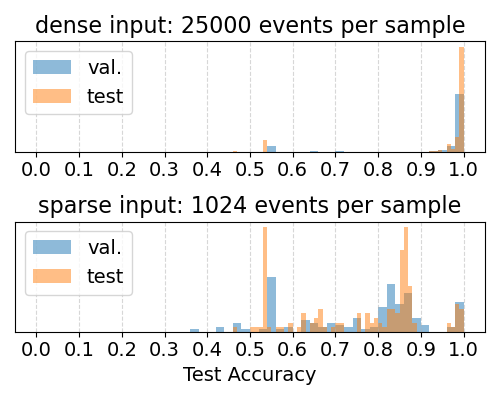}} & \includegraphics[width=30mm]{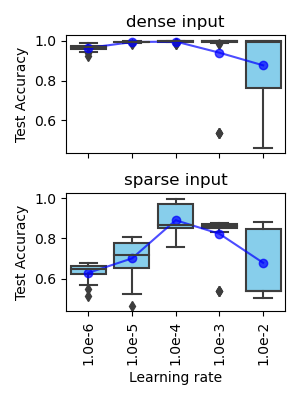} & \includegraphics[width=30mm]{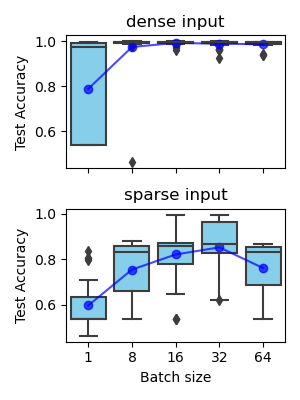} & \includegraphics[width=30mm]{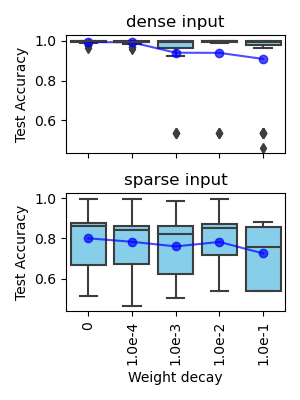} \\
\rotatebox{90}{\scriptsize DVS-Gesture} & {\includegraphics[width=50mm]{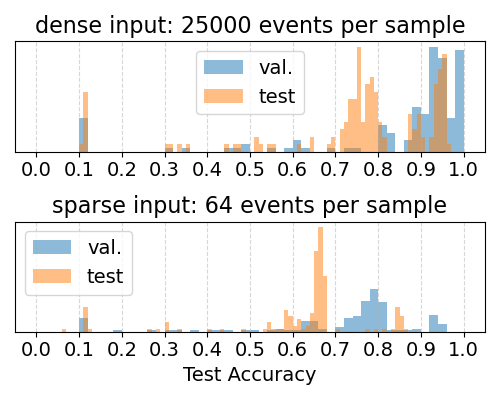}} & \includegraphics[width=30mm]{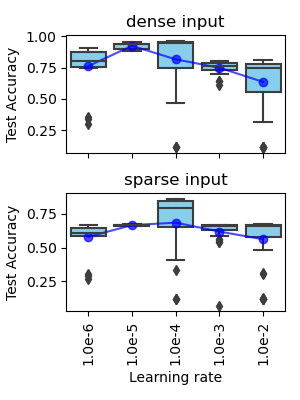} & \includegraphics[width=30mm]{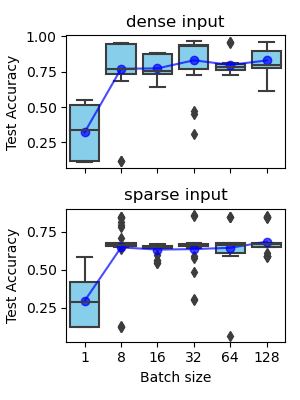} & \includegraphics[width=30mm]{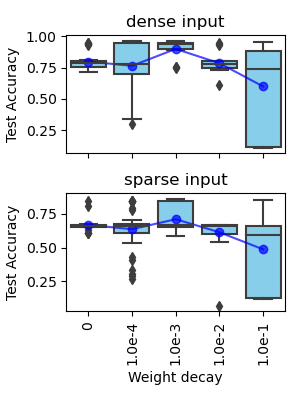} \\
\rotatebox{90}{\scriptsize N-ASL} & {\includegraphics[width=50mm]{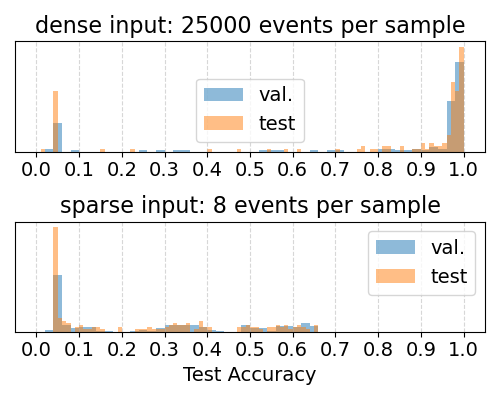}} & \includegraphics[width=30mm]{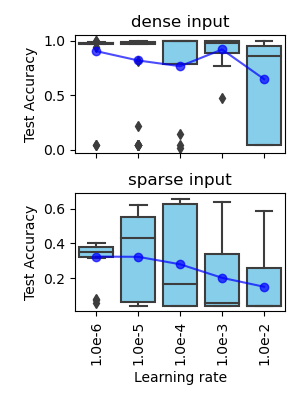} & \includegraphics[width=30mm]{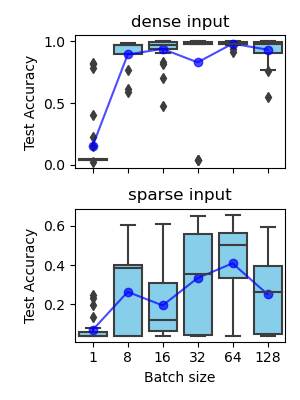} & \includegraphics[width=30mm]{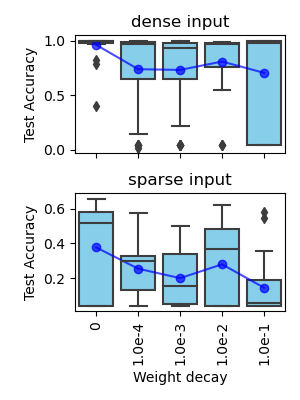} \\ 
\rotatebox{90}{\scriptsize N-Caltech101} & {\includegraphics[width=50mm]{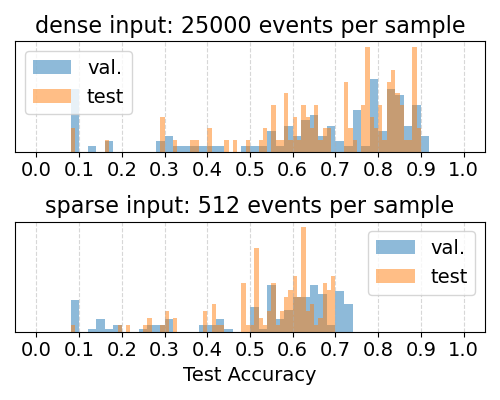}} & \includegraphics[width=30mm]{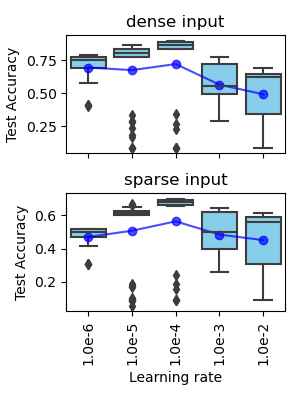} & \includegraphics[width=30mm]{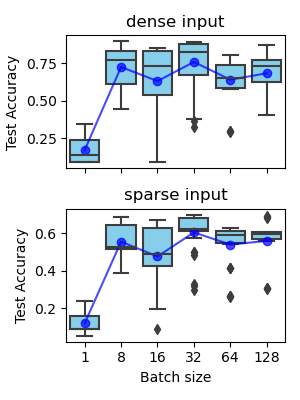} &  \includegraphics[width=30mm]{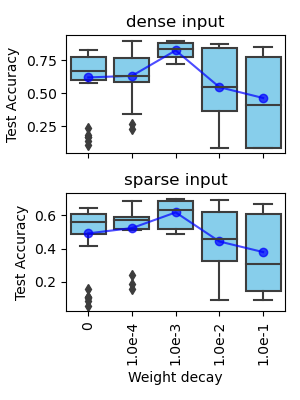}\\ 
\bottomrule 
\end{tabular}}
\caption{%
Histograms of all 300 accuracies attained during hyperparameter tuning for both dense and sparse input scenarios (first column), and boxplots of test accuracies obtained using each individual hyperparameter: learning rate, batch size, and weight decay (second to fourth columns).
The blue curve represents the mean test accuracies.%
}
\label{fig:HP histograms}
\end{figure}

\Cref{fig:HP histograms} illustrates histograms of accuracies for all 300 experiments with different HP sets. 
For datasets except N-Caltech101, we observe a notable distinction between the dense and sparse cases.
In the dense case, runs for different HP sets tend to cluster near the highest achievable accuracy, while in the sparse case, we observe a noticeable gap between the main cluster of runs and those achieving the highest accuracy. 
% This suggests that in sparse scenarios, hyperparameter tuning runs may often fall within a range of accuracy that is not optimal, indicating a higher sensitivity to hyperparameters.
% Conversely, in dense cases, performance is more likely to be close to the optimal accuracy without extensive HP tuning.

% \newcolumntype{M}[1]{>{\centering\arraybackslash}m{#1}}

\clearpage
\Cref{fig:HP histograms} also shows the dependencies of test accuracies to each HP individually.
Each box in the boxplot figures of the second to fourth columns shows the test accuracies corresponding to all the experiments with specific values of that individual hyperparameter
Notably, in the sparse case, sensitivity to individual HPs is higher compared to the dense case. 
Among the HPs, the variation of learning rate more significantly affects the test accuracies.
% It's important to note that our focus is not on comparing the maximum achievable accuracies between the dense and sparse cases.
% Instead, we emphasize how closely most runs are near the maximum achievable accuracies. 
The behavior of the N-Caltech101 dataset differs from that of other datasets, suggesting that sensitivity to HPs is not a universal behavior in sparse input cases. 
Nevertheless, we argue that such sensitivity may arise in some datasets, and propose to consider extensive hyperparameter tuning during model development in sparser scenarios.
In the N-Caltech101 dataset, it appears that the loss of information plays a more significant role in the accuracy drop than sensitivity to HPs.
% One possible explanation for this behavior is the classes are highly dependent on spatial data than the temporal data. 

\Cref{fig:HP tuning strings} presents parallel coordinate plots of hyperparameter (HP) tuning for the Fan1vs3 dataset using dense and sparse inputs from the Weight and Biases \cite{wandb} website.
Each string represents a set of HPs, and we observe that in the dense case, the concentration of test accuracies is near the peak. Conversely, in the sparse case, only a small number of HP sets manage to achieve accuracy levels close to the peak.
This suggests the increased importance of HP tuning during training in sparse cases to ensure that performance reaches its maximum potential.

To quantify the level of sensitivity to HPs, we propose a \emph{hyperparameter sensitivity metric} based on the clustering of accuracies for different HP sets.
First, we employ the $K$-means algorithm to cluster the accuracies into $K$ clusters.
Next, we compute the distance between the centers of the most populated cluster (i.e., the cluster with the most runs) and the cluster with the maximum accuracy. 
To account for the different maximum accuracies in dense and sparse cases, we normalize the obtained distance by dividing it by the maximum achieved accuracy.
In scenarios where the number of runs concentrated near the maximum accuracies is low, such as in the Fan1vs3 dataset, we need to select a larger number of $K$ to prevent the high-accuracy cluster from merging with other clusters incorrectly.
For this, we compute the normalized metric for a range of $K$ values, i.e. from 2 to 10.
Then, we select the maximum value over different $K$ values in search of at least one popular cluster falling far apart from the high-accuracy cluster.
\Cref{tab:HP_sensitivity_metric} displays the HP sensitivity metric, as well as the mean and maximum test accuracy and the percentage of improvement observed from mean to maximum accuracy for each dataset for dense and sparse cases.
% Notably, we observe a larger gap between the popular cluster and the cluster with maximum accuracies in sparse cases, indicating a higher sensitivity to hyperparameters when the input is sparse.
In most datasets, excluding N-Caltech101, the metric indicates a higher sensitivity to HPs when the input is sparse, and the improvement from mean to maximum accuracy is more substantial in the sparse case.

\begin{table}[tb]
% \begin{wraptable}{thr}{0.5\textwidth}
    \centering
    \caption{%
Hyperparameter (HP) sensitivity metric along with mean and maximum test accuracies for dense and sparse cases for different datasets. Higher values for the metric $\uparrow$ indicate a higher sensitivity to HPs.
For most datasets, the HP sensitivity metric as well as the improvement from mean to maximum test accuracy is larger in the sparse case (in bold) compared to the dense case.}
    \label{tab:HP_sensitivity_metric}
    \resizebox{1\textwidth}{!}{%
    \begin{tabular}{lccc@{\hskip 0.2in}ccc@{\hskip 0.2in}ccc@{\hskip 0.2in}cc}
\toprule
Dataset & \multicolumn{2}{c}{N-ASL} & & \multicolumn{2}{c}{Fan1vs3} & & \multicolumn{2}{c}{DVS-Gesture} & & \multicolumn{2}{c}{N-Caltech101}\\\cmidrule{2-3} \cmidrule{5-6} \cmidrule{8-9} \cmidrule{11-12}
 & \small{dense} & \small{sparse} & & \small{dense} & \small{sparse} & & \small{dense} & \small{sparse} & & \small{dense} & \small{sparse}\\
\small{\# events} & \small{25000} & \small{8} & & \small{25000} & \small{1024} & & \small{25000} & \small{64} & & \small{25000} & \small{512}\\
\midrule
\multicolumn{12}{c}{\textbf{Hyperparameter sensitivity metric}}\\
\midrule
val. metric & 0.000 & \textbf{0.915} & & 0.000 & \textbf{0.471} & & 0.060 & \textbf{0.191} & & \textbf{0.122} & 0.098\\
test metric & 0.021 & \textbf{0.887} & & 0.000 & \textbf{0.145} & & 0.196 & \textbf{0.230} & & \textbf{0.279} & 0.101\\
\midrule
\multicolumn{12}{c}{\textbf{Mean and maximum test acc.}}\\
\midrule
mean test acc. (\%) & 80.74 & 25.09 & & 95.14 & 76.62 & & 75.22 & 61.50 & & 69.23 & 55.41\\
max. test acc. (\%) & 99.95 & 65.57 & & 100.00 & 99.62 & & 96.37 & 86.34 & & 89.55 & 69.88\\
max. to mean improvement (\%) & 23.79 & \textbf{161.34} & & 5.10 & \textbf{30.01} & & 28.13 & \textbf{40.38} & & \textbf{29.34} & 26.11\\
\bottomrule
\end{tabular}
}
% \end{wraptable}
\end{table}

\subsection{Gradient Diversity in the Sparse vs. Dense Case}
\label{sec:experiments:subsec:grad_div}

In this subsection, we examine how the subsampling level of an event video affects the gradients in the network. 
We first randomly select a video from the training set of the Fan1vs3 dataset and generate $M=100$ different subsamples from it for both the sparse (1,024 events) and the dense (25,000 events) case.
For each subsample, we compute the loss and obtain the gradient of the loss with respect to the weights of $L=5$ different layers, corresponding to the last convolutional layer of each block in the ResNet34 network. 
We pick these layers to study the gradients at different depths of the network.
The specific layers chosen for gradient computation are detailed in the supplementary material. %TODO: Supp
For subsample  $i\in\{1,\ldots,M\}$ and layer $l\in\{1,\ldots,L\}$, we denote the gradient vector as $v_i^{(l)}\in\mathbb{R}^{p_l}$, where $p_l$ is the number of learnable weights in layer $l$.
Then, we compute the pairwise cosine similarity between $v^{(l)}_i$ and $v^{(l)}_j$ for all unique pairs in $\{(i,j)| 1\le i < j \le M\}$ for each layer.
The total number of cosine similarities for each layer equals $\frac{M\,(M-1)}{2} = 4950$.
To demonstrate the diversity of the gradients, we illustrate the histogram of the cosine similarities separately for each layer for both sparse and dense cases.

\begin{figure}[tb]
    \begin{subfigure}[b]{1\textwidth}
\includegraphics[width=1\textwidth]{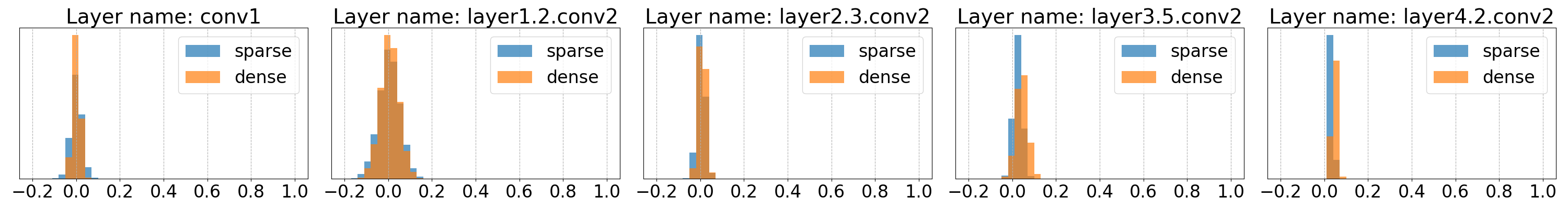}
        \caption{Untrained network}
        \label{fig:grad_diversity_untrained}
    \end{subfigure}
    \\
\begin{subfigure}[b]{\textwidth}
\includegraphics[width=1\textwidth]{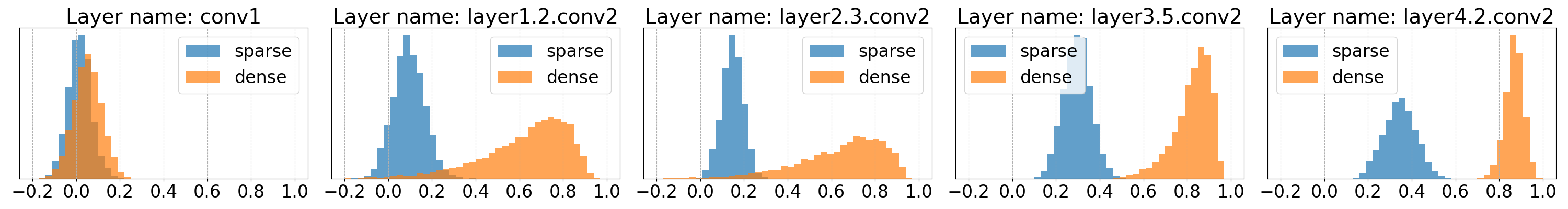}
        \caption{Trained network}
        \label{fig:grad_diversity_fully_trained}
\end{subfigure}
    \caption{Histogram of cosine similarities for both dense and sparse inputs for the (a) untrained network and (b) trained network. Values close to 1 indicate greater alignment among gradients. For the trained network, gradients are more aligned in the dense case, and as we progress to deeper layers (from left to right), the alignments increase for both sparse and dense cases.}
    \label{fig:histogram_of_grads}
\end{figure}

\Cref{fig:histogram_of_grads} displays the histograms of cosine similarity between unique pairs of gradient vectors. 
We show histograms for both untrained and fully trained networks. %
 Values closer to 1 in the histograms show greater alignment among the gradients, while values close to 0 show higher gradient diversity.
 At the beginning of training, when the network is untrained, the gradients are diverse for both sparse and dense cases.
For the trained network, we observe that gradients are more aligned in the dense case compared to the sparse case, which is in line with the results of \cref{sec:experiments subsec: HPO} about the difficulty of training networks with sparse inputs.
% This result aligns with the expectation that sparse inputs introduce more variability and uncertainty into the training process, complicating the convergence of the network towards optimal weight values.
 Higher gradient diversity suggests that the network may not travel far from the initial weights during gradient descent, instead taking a random walk around the initialization point.
% higher uncertainty the network has in deciding to classify the input video. 
% \textbf{Nergis: I don't understand this sentence at all. Also, how can we relate gradients in different layers, e.g. a middle layer, to the decisions? Do we have higher uncertainty in the early layers?
% I would maybe phrase this differently/more literally: Something like higher gradient diversity and the network will not travel far from the initial weights during gradient descent, but travel in a random walk around the initialization point. But maybe you can find something smarter to say.}
Conversely, greater consensus among the gradients of different subsamples during training, i.e. closer values to 1, contributes to faster convergence of the network towards final weight values.
Furthermore, as we progress to deeper layers of the trained network, gradients tend to be more aligned for both sparse and dense inputs.
This can be attributed to the larger receptive field in deeper layers, allowing the network to develop a global understanding of the input video, thereby reducing the influence of different subsampling on these layers.

\section{Discussion and Conclusion}
\label{sec:discussion}

In this paper, we study the effect of subsampling on the sparsity vs. accuracy trade-off in event video classification using a CNN model.
Our findings reveal that accuracy can remain high even under severe sparsity in many datasets. 
However, special attention must be given to the training challenges arising from hyperparameter (HP) sensitivity in sparse input regimes.

\subsubsection{Practical Implications for the Subsampling Procedure}
The training procedure described in \cref{sec:method:subsec:subsampling_procedure}, where a new subset of events is selected in each epoch, may not be feasible in all scenarios. 
It is important to note that when dense data is available, this subsampling procedure can help mitigate overfitting. 
In practice, data collection for training occurs less frequently than in the utilization phase of the model, and in the former case, we can leverage higher sampling rates using capturing devices with greater bandwidth or computational capacity. 
These dense datasets can then be used to train the models following our procedure, helping the model maintain higher accuracies for sparse event input at test time.

\subsubsection{Limitations} 
It is important to note that keeping accuracies high with sparse input is not universal across all datasets. 
As a specific example, we observe that in the N-Caltech101 dataset, accuracies drop faster with increased subsampling, and, unlike in other datasets, hyperparameter tuning does not help much in the sparse case compared to the dense case. 
This dataset is different from others we study because it is derived from static images captured by moving the camera, unlike the others, which are recordings of real-world scenes. 
The high sensitivity to sampling level in N-Caltech101 likely stems from its greater reliance on spatial information rather than temporal information.

The results of this paper rely on the subsampling procedure of selecting random subsets of events at each epoch. 
However, dense event data may not always be available to subsample from during the training process. 
Therefore, maintaining high accuracies may not be achievable, and the model is prone to overfitting if the training set itself is sparse, and limited in sample size. 

\subsubsection{Future Work} 
We limited our analysis to classification using CNNs as a proof-of-concept.
However, our training methodology simply allows for increasing the sparsity in favor of reducing the bandwidth at train and test time, which might be a viable training procedure for other popular event-based vision models, including transformers and graph neural networks (GNNs).
Similarly, looking at the gradient analysis, the instabilities seem to emerge, not due to the nature of the CNNs or the classification task, but simply due to the variability of information in the training set in the sparse setting. 
This suggests that our results could potentially be generalized to other models and vision tasks, making it a future direction for empirical testing.
Additionally, adapting methods such as \cite{uhrig_sparsity_2017}, which aim to mitigate the impact of input sparsity in conventional images, can be an interesting line of research to improve the robustness of CNNs to sparse event data.

% \subsubsection{Broader impact}
% This work has broader implications for enhancing the efficiency of machine vision by demonstrating that high accuracy can be maintained even with significantly reduced event data. 
% The results of this study contribute to the general goal of reducing computational and bandwidth costs, aligning with the direction of more environmentally friendly AI practices for edge devices.

\bibliographystyle{splncs04}
\bibliography{other_refs,references_tex}

\title{Supplementary Material for the Paper: Pushing the boundaries of event subsampling in event-based video classification using CNNs} 

\titlerunning{Supplementary Material}

\author{Hesam Araghi\orcidlink{0000-0002-4539-4408} \and
Jan van Gemert\orcidlink{0000-0002-3913-2786} \and
Nergis Tomen\orcidlink{0000-0003-3916-1859}}

\authorrunning{H.~Araghi et al.}

\institute{
Computer Vision Lab, Delft University of Technology\\
\email{\{h.araghi, j.c.vangemert, n.tomen\}@tudelft.nl}}

\maketitle

\section{Details of Training Procedures}
\label{sec:supp:training details}

For all experiments, we use Adam \cite{kingma_adam_2017} for optimization.
The learning scheduler is the `Reduce on Plateau' with a reduction factor of 2, and it monitors the validation accuracy.
The cross-entropy loss function is used for the classification.
During testing, the evaluation is conducted 20 times with different random subsamples for the test set, and the average accuracy is reported. 
We use PyTorch and PyTorch Lightning libraries for the training procedure and use Weights and Biases for logging the results \cite{wandb}.

\subsection{EST Algorithm \cite{gehrig_end--end_2019} Details} 
We use the implementation code provided in the original paper's GitHub Repository. %
\footnote{\href{https://github.com/uzh-rpg/rpg_event_representation_learning}{https://github.com/uzh-rpg/rpg\_event\_representation\_learning}}
For all experiments, we choose $C = 9$ frames (addressed as the number of bins in the original paper) for each positive and negative polarity, yielding a total of 18 channels.
As the CNN classifier, we use a ResNet34~\cite{he_deep_2015}
model pre-trained on ImageNet1k\_v1, modifying the number of input channels from 3 to 18.
For the filter function  $f_{\theta}(\cdot): \mathbb{R}\rightarrow\mathbb{R}$, we employed an MLP model with two hidden layers, each containing 30 nodes, and used Leaky ReLU with a negative slope of $0.1$ as the activation function.

\subsection{Details for `Accuracy vs. Sparsity in Event Data Classification' Experiment}

\Cref{table:supp:acc_vs_sparsity_HPs} lists the hyperparameters used for the experiment `Accuracy vs. Sparsity in Event Data Classification'.
The hyperparameters are chosen to be the same across all sparsities of the same dataset.
 For each dataset, we picked a set of hyperparameters which could converge to a high accuracy at every sparsity level based on limited preliminary experiments. 
 For N-Cars, we adopt the values from the EST paper \cite{gehrig_end--end_2019}.

 \begin{table}[tb]
\centering
\caption{Hyperparameters used for `Accuracy vs. Sparsity in Event Data Classification' experiment}
\resizebox{\textwidth}{!}{%
\begin{tabular}{lccccc}
\toprule
\textbf{Hyperparameters} & \textbf{N-ASL} & \textbf{N-Caltech101} & \textbf{N-Cars} & \textbf{DVS-Gesture} & \textbf{Fan1vs3} \\ 
\midrule
Learning Rate & 1e-4 & 1e-4 & 1e-5 & 1e-4 & 1e-4 \\ 
Patients for Learning Scheduler & 20 & 40 & 75 & 40 & 40 \\ 
Weight Decay & 0 & 0 & 0 & 0 & 0 \\ 
Number of Epochs & 100 & 250 & 500 & 250 & 250 \\ 
Batch Size & 32 & 16 & 100 & 64 & 32 \\ 
\bottomrule
\end{tabular}}
\label{table:supp:acc_vs_sparsity_HPs}
\end{table}

\section{Accuracy curves for Table~1 of the paper}
\label{sec:supp:accuracy curves}

\begin{figure}[tb]
    \centering
    \includegraphics[width=\linewidth]{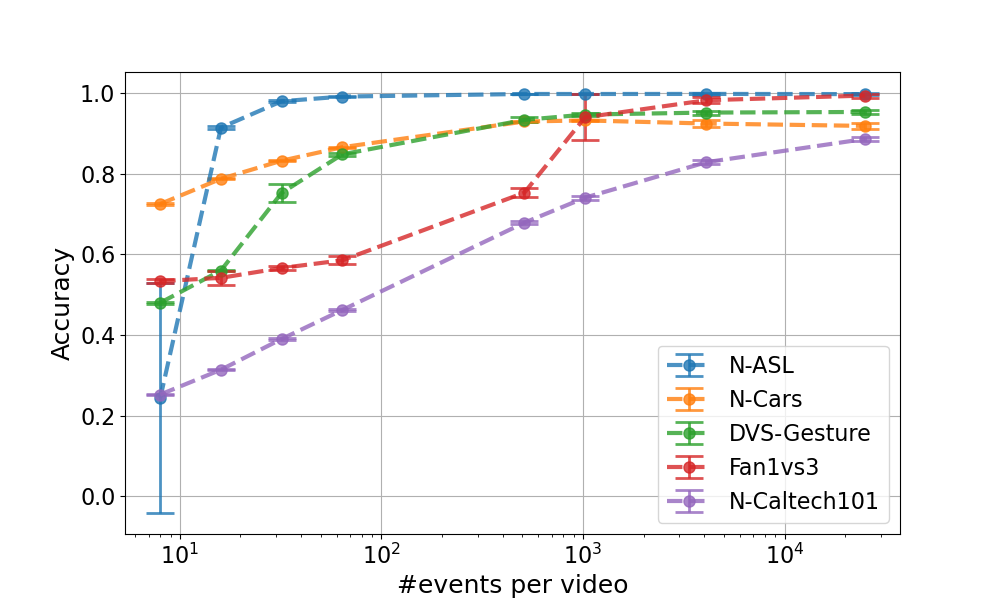}
    \caption{Classification accuracy curves as the number of events per video decreases across various datasets. The error bars represent the standard deviation of accuracies across different runs. For many datasets, the accuracy curves do not significantly drop compared to the dense input case.}
    \label{fig:supp:accuracy curves}
\end{figure}

For better visualization of the accuracy behavior for the number of events per video, the accuracy curves for different datasets are depicted in \cref{fig:supp:accuracy curves}. 
Except for the N-Caltech101 dataset, the accuracies remain relatively close to the dense case (rightmost points). 
The curves remain nearly flat until they reach a point where the accuracy drops significantly.

\section{Details of Event Classification Datasets}
\label{sec:supp:datsets details}

% TODO: Samples of each in supp material (or only Fan dataset)
%TODO:  train val test split for each dataset
%TODO: image resolutions
% TODO: table with number of events and classes and number of videos

\Cref{table:supp:datasets_characteristics} provides the characteristics of the different datasets we used for event classification in this paper. 
For splitting the data into training, validation, and test sets, we used the following approach: for datasets that do not come with a predefined test set, we randomly divided the videos of each class into 75\% for training, 10\% for validation, and 15\% for testing. 
For datasets that already include a test set, we randomly divided the training set videos into 85\% for training and 15\% for validation.

For the DVS-Gesture dataset \cite{amir_low_2017}, we followed a procedure similar to \cite{subramoney_efficient_2023}. 
We used the Tonic library \cite{lenz_tonic_2021} to download and load the dataset. 
The videos in the dataset were segmented by time into videos of length 1.7 seconds, without any overlap. 

 \begin{table}[tb]
\centering
\caption{Characteristics for different event classification datasets used in this paper}
{%
\begin{tabular}{lccccc}
\toprule
\textbf{Characteristics} & \textbf{N-ASL} & \textbf{N-Caltech101} & \textbf{N-Cars} & \textbf{DVS-Gesture} & \textbf{Fan1vs3} \\ 
\midrule
\# classes          & 24    & 101   & 2     & 11    & 2 \\ 
\# videos           & 100,800    & 8,709    & 24,029    & 1,342    & 510 \\ 
spatial resolution  & [240,180]     & [240,180]     & [100,120]     & [128,128]     & [1280,720] \\ 
separate test set   & \ding{55}   & \ding{55}   & \ding{51}   &  \ding{51}  & \ding{55} \\
\bottomrule
\end{tabular}}
\label{table:supp:datasets_characteristics}
\end{table}

\subsubsection{Fan1vs3 dataset} 
We introduce a new toy dataset designed to emphasize the temporal details captured by event cameras. To create this dataset, we placed a fan in front of a Prophesee Gen4 event camera \cite{finateu_a_1280_2020}. The fan blades were set to two different speeds: the slowest (speed 1) and the fastest (speed 3), corresponding to the two classes of the dataset. 

\begin{figure}[tb]
    \centering
    \includegraphics[width=0.8\linewidth]{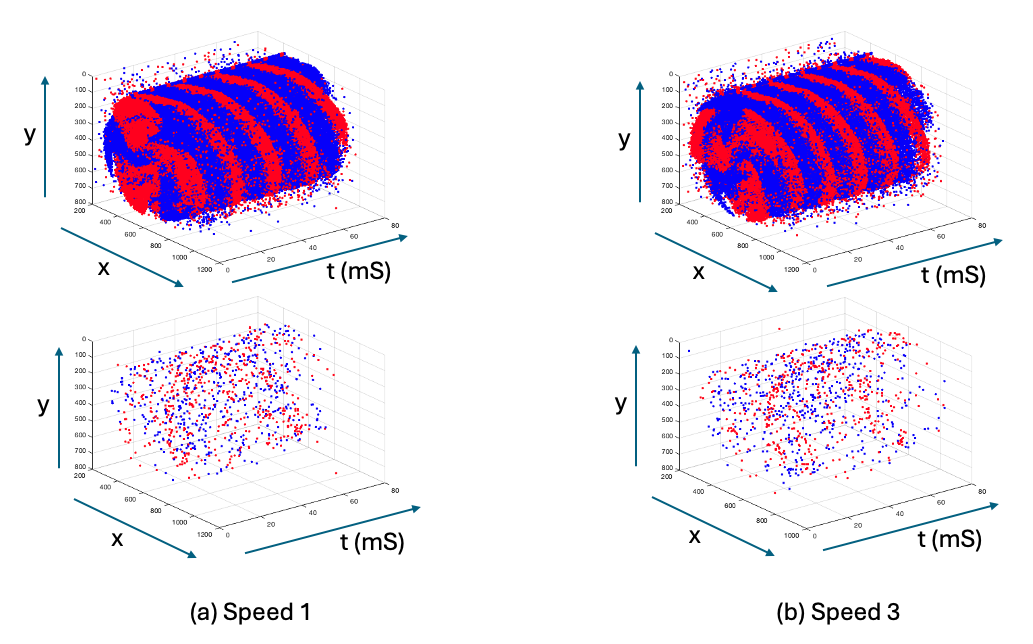}
    \caption{Event illustration of a video for each speed class of Fan1vs3 dataset in spatiotemporal space. The second row displays the subsampled events, reduced to 1024 events.}
    \label{fig:fan1vs3_events}
\end{figure}

The videos for each class are segmented into 75-millisecond clips. The `speed 1' class contains 235 videos, while the `speed 3' class contains 275 videos. \Cref{fig:fan1vs3_events} displays the events of an example video for each speed class in spatiotemporal space, along with the subsampled version of each, reduced to 1024 events. This dataset allows us to investigate the classification with an emphasis on the temporal data.

% \begin{figure}[t]
%     \centering
%     \includegraphics[width=0.4\linewidth]{images/appendix_images/fan_pic.jpg}
%     \caption{The fan used for creating the toy dataset. The event camera recorded the rotating blades of the fan with two classes: 1. Speed 1 (fan's slowest speed) and 2. Speed 3 (fan's fastest speed). The two classes are the speed buttons on the fan.}
%     \label{fig:fan_pic}
% \end{figure}

\section{Details for `Gradient Diversity in the Sparse vs. Dense Case' Experiment}
\label{sec:supp:grad layers details}

\begin{figure}[tb]
    \centering
    \includegraphics[width=1\linewidth]{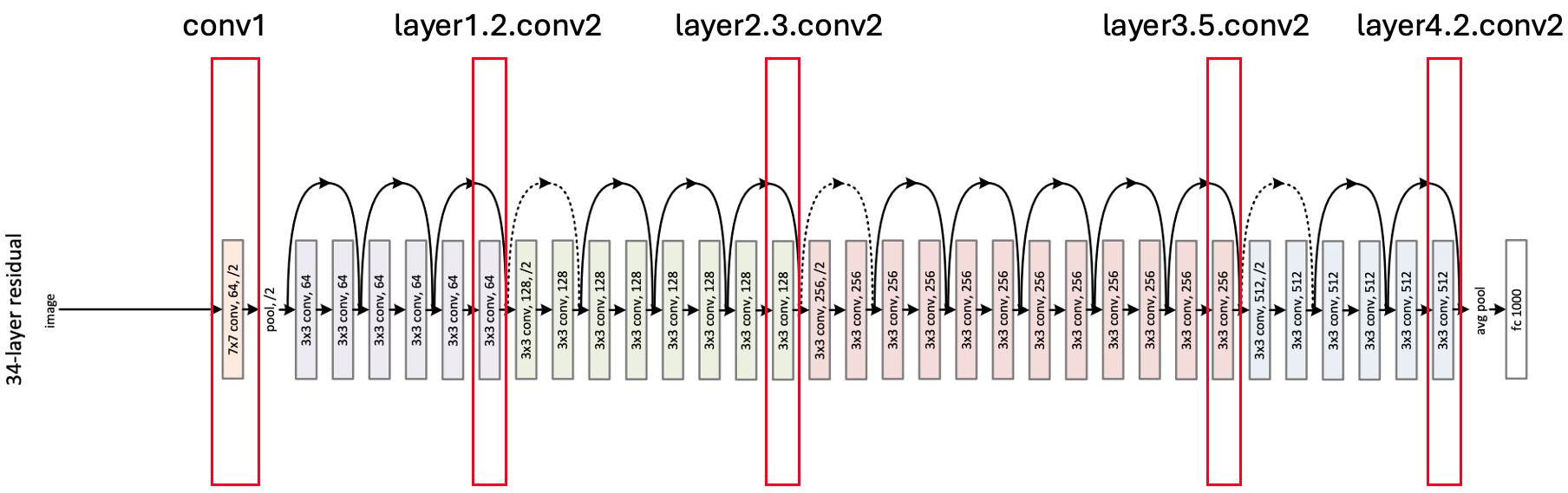}
    \caption{Layer of the ResNet34 network which is used in gradient analysis of the `Gradient Diversity in the Sparse vs. Dense Case' Experiment. The background image is adapted from the original ResNet paper \cite{he_deep_2015}.}
    \label{fig:resnet_layers}
\end{figure}

For computing the gradients in the experiment `Gradient Diversity in the Sparse vs. Dense Case', we consider five different layers, specifically the last convolutional layer of each block in the ResNet34 network. 
The specific names of these convolutional layers along with the number of parameters for each layer are provided in \cref{tab:supp_grad_layers_name}. 
\Cref{fig:resnet_layers} also illustrates the position of the selected layer in the network architecture.

\begin{table}[tb]
    \centering
    \caption{Layer names and the number of parameters in each layer used in gradient analysis of `Gradient Diversity in the Sparse vs. Dense Case' Experiment.}
    \begin{tabular}{cc}
    \toprule
      Layer name  & \# parameters \\
      \midrule
       \verb+conv1+ & 9,408 \\
       \verb+layer1.2.conv2+ & 36,864 \\
       \verb+layer2.3.conv2+  & 147,456 \\
       \verb+layer3.5.conv2+ & 589,824 \\
       \verb+layer4.2.conv2+ & 2,359,296 \\
       \bottomrule
    \end{tabular}
    \label{tab:supp_grad_layers_name}
\end{table}

\section{Details of Compute Resources}
\label{sec:supp:compute resources}

For the computing resources, we used a cluster of machines with GPUs,  which it will be referenced later. 
We used one GPU per training execution. 
The GPUs used for the experiments were primarily NVIDIA A40 or NVIDIA V100 models. 
Each job requested 16 GB of RAM and 4 CPU cores. 
The specific computational times for the experiments in the `Hyperparameter Sensitivity with Increasing Sparsity' subsection, which involved evaluating 300 hyperparameter sets per dataset, are detailed in \cref{tab:supp:GPU days}. 
The computational times are reported from the Weight and Biases \cite{wandb} website used for logging the results.

\begin{table}[tb]
    \centering
    \caption{GPU computation time on the cluster in GPU days for the hyperparameter tuning used in the `Hyperparameter Sensitivity with Increasing Sparsity' experiment.}
    \begin{tabular}{cc}
        \toprule
        Dataset & GPU computation time (GPU days)\\
        \midrule
        N-ASL & 280\\
        N-Caltech101 & 67\\
        DVS-Gesture & 24\\
        Fan1vs3 & 7\\
        \midrule
    \end{tabular}
    \label{tab:supp:GPU days}
\end{table}

\clearpage


\end{document}